\documentclass[runningheads]{llncs}
\usepackage{esvect}
\usepackage[T1]{fontenc}
\usepackage{graphicx,verbatim}
\usepackage{amsfonts}
\usepackage{booktabs}
\usepackage{multirow}
\usepackage{amsmath}
\usepackage{marvosym}
\begin{document}
\title{Cellular-Communication-Level Interpretability for Pathology Foundation Models via Graph Distillation on Microenvironment}
\titlerunning{Interpretable Pathology Foundation Model via Graph Distillation}
%

\author{
Yuxiang Xiao\inst{1}\and
Zhiwei Chen\inst{1}\and
Dan Dai\inst{5}\and
Wei Li\inst{6} \and
Tianyang Zhang\inst{4} \and
Yakun Ju\inst{2} \and
Yang Hu\inst{2,3,4}\textsuperscript{(\Letter)} \thanks{Yang Hu and Kaixiang Yang contribute equally to this work.\\
Corresponding email: \email{superhy199148@hotmail.com} and
\email{yangkx@scut.edu.cn}} \and
Kaixiang Yang\inst{1}\textsuperscript{(\Letter)}$^{\star}$
}

\authorrunning{Yuxiang Xiao, Zhiwei Chen et al.}

\institute{
School of Computer Science and Engineering, South China University of Technology, China
\and
School of Computing and Mathematical Sciences, University of Leicester, UK
\and
Leicester Cancer Research Centre, University of Leicester, UK
\and
Department of Engineering Science, University of Oxford, UK
\and
School Of Computer Sci And Digital Tech, Aston University, UK
\and
ZoyMed, China
}

\maketitle              
\begin{abstract}
Pathology foundation models (PFMs) provide strong tile-level representations but remain difficult to interpret at the cellular and microenvironmental scales that underpin clinical reasoning. We introduce Graph-Interpreter (\textbf{G-Interp}), a graph-distillation framework that equips a frozen PFM teacher with a cellular-communication-level “plug-in” interpreter, without modifying the teacher. For each tile, we segment cells as graph nodes and construct a microenvironment graph based on spatial adjacency. Graph neural network (GNN) students distil the PFM embedding, whilst learning attention-based message passing that yields node- and edge-level importances. We interpret these importances as cell–cell communication evidence, providing fine-grained explanations of how PFMs encode microenvironmental context.
To stabilise distillation when graph abstraction is imperfect, we employ a lightweight auxiliary student to supply complementary visual cues and condition graph message passing, while keeping the primary interpretability signal graph-derived. We evaluate explanation faithfulness by mapping graph-selected evidence back to the image using instance masks and measuring teacher sensitivity under targeted \textit{vs} non-target occlusions. Across multiple histopathology tasks, \textbf{G-Interp} produces highly scalable, microenvironment-aware explanations, while maintaining competitive predictive performance.

\keywords{Pathology foundation models  \and Graph distillation \and Microenvironment graphs \and Explainable AI.}

\end{abstract}
\section{Introduction}

Pathology Foundation Models (PFMs) have advanced computational pathology by providing pre-trained encoders that yield robust representations for diverse downstream tasks \cite{li2025survey}. However, two challenges hinder clinical translation. First, their scale (often hundreds of millions of parameters) incurs substantial computational cost, complicating deployment on resource-constrained clinical terminals, especially for Whole Slide Images (WSIs) \cite{song2023artificial}. Second, their black-box representations conflict with histopathological practice, which demands cellular-level evidence and microenvironment context \cite{tosun2020explainable,ali2025graph,hu2025self}.

Knowledge distillation (KD) \cite{hinton2015distilling} is widely used to compress large models into efficient architectures \cite{burhan2024comprehensive}. Yet most pathology distillation treats the student as a pure mimic, aligning visual feature maps between a large teacher and a smaller vision-based student. While effective at transferring \emph{visual appearance cues} (e.g., stain- and morphology-related patterns), such matching rarely encourages the student to preserve or expose structured biological priors (e.g., cellular topology). An important next step is therefore to distill the teacher's latent understanding of the cellular ecosystem into a compact and interpretable structure.

To this end, we propose \textbf{G-Interp}, a graph-visual distillation framework that adapts frozen PFMs into a lightweight, interpretable student. G-Interp adopts a dual-branch student: a \textit{Graph Branch} that constructs a patch-level cell graph using NuLite \cite{TOMMASINO2026109333} and encodes it with a lightweight GNN, and a \textit{Visual Branch} (ResNet-18) that captures global tissue morphology. The two branches are fused via a gating mechanism, enabling adaptive balancing between cellular evidence and global context. Importantly, the graph branch provides explicit cell- and interaction-level explanations via its message-passing weights.

Our main contributions are:

(1) We propose \textbf{G-Interp}, a graph-visual dual-branch distillation framework that transfers knowledge from large, opaque PFMs to lightweight students, achieving strong predictive performance under multiple fusion strategies.

(2) We introduce a microenvironment graph branch as a transparent proxy that translates abstract PFM representations into explicit cellular nodes and topological interactions, enabling biologically interpretable explanations.

(3) On three public breast cancer datasets, we validate both performance and interpretability, showing that G-Interp preserves structural biological priors and yields clinically meaningful microenvironmental evidence.

\section{Related Work}

\noindent\textbf{Pathology Foundation Models.}
Self-Supervised Learning (SSL) has driven large-scale pre-training on millions of histopathology tiles. ViT-based PFMs such as Phikon v2 \cite{filiot2024phikon}, UNI v2 \cite{chen2024towards}, and Virchow2 \cite{zimmermann2024virchow2} provide general-purpose representations for diverse tasks, but their computational footprint hinders routine deployment \cite{li2024rethinking}. Although ViTs can produce attention maps (e.g., rollout), these are typically coarse and do not resolve cell boundaries or directional cell--cell interactions \cite{xu2023vision}.

\noindent\textbf{Knowledge Distillation in Computational Pathology.}
KD transfers a large teacher's generalisation to a compact student. In pathology, where a WSI may contain tens of thousands of patches, faster inference is desirable \cite{wang2022lymph,wang2024histo}. Most work aligns dense teacher embeddings to a smaller vision student (e.g., ResNet), but distance-based matching remains a black box and does not yield biologically meaningful entities. G-Interp instead distils into a graph-structured student to expose cell-level importance and interaction pathways, linking compression with clinical transparency.

\noindent\textbf{Graph Neural Networks for Microenvironment Modelling.}
GNNs encode cells (or regions) as nodes and spatial relations as edges, and are widely used for survival prediction, subtyping, and grading \cite{fu2023application,gogoshin2023graph,song2023artificial,waqas2024multimodal}. Conventional pipelines train GNNs from scratch (fully supervised or MIL), which can be sensitive to noisy WSI labels and unstable optimisation \cite{shi2025slide,lin2026sg}. In contrast, G-Interp trains the GNN as a distillation student from dense PFM representations, providing both microenvironment features and cellular-level explanations.

\section{Method}

\begin{figure*}[!t]
    \centering
    \includegraphics[width=\textwidth]{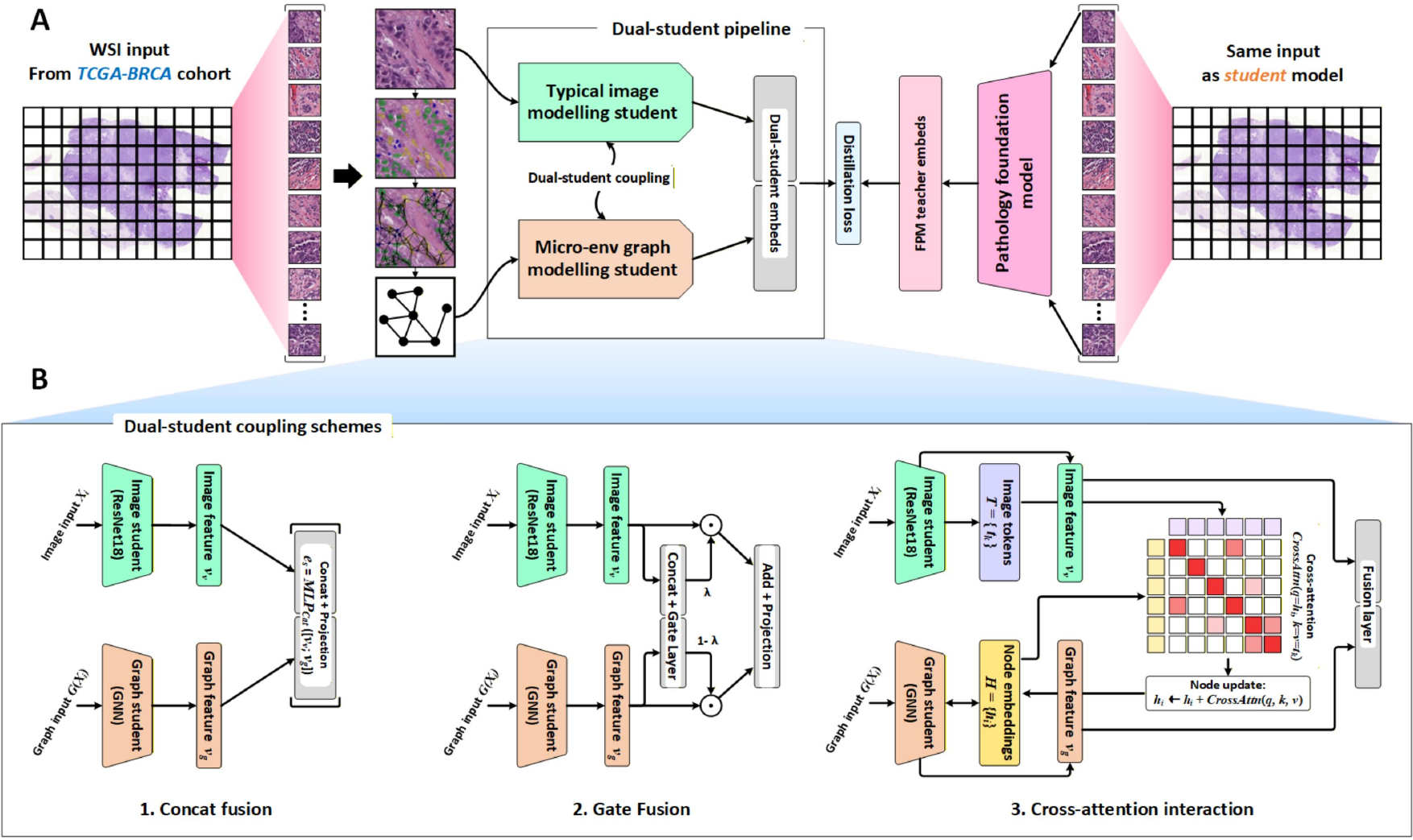} 
    \caption{\textbf{Overview of G-Interp.} A frozen PFM teacher provides target embeddings for each cell-containing tile. The dual-branch student processes the tile in parallel: a Visual Branch (ResNet-18) captures global tissue appearance, and a Graph Branch builds and encodes a microenvironment graph from NuLite segmentations. The fused features are optimised to match the teacher embedding.}
    \label{fig:framework}
\end{figure*}

\subsection{Overall Distillation Framework}
As illustrated in Fig.~\ref{fig:framework}, G-Interp distils representations from a frozen PFM teacher into a lightweight dual-branch student. Given an image tile $\mathbf{X}$, the teacher outputs a high-dimensional embedding $\mathbf{e}_T \in \mathbb{R}^{D}$, and the student produces a corresponding embedding $\mathbf{e}_S \in \mathbb{R}^{D}$. Following our implementation, we first $\ell_2$-normalise the embeddings, $\hat{\mathbf{e}}_S=\mathbf{e}_S/\|\mathbf{e}_S\|_2$ and $\hat{\mathbf{e}}_T=\mathbf{e}_T/\|\mathbf{e}_T\|_2$, and then minimise a cosine embedding loss with a positive target label $y=1$:
\begin{equation}
    \mathcal{L}_{\mathrm{KD}} = 1 - \cos(\hat{\mathbf{e}}_S, \hat{\mathbf{e}}_T)
    = 1 - \frac{\hat{\mathbf{e}}_S^{\top}\hat{\mathbf{e}}_T}{\|\hat{\mathbf{e}}_S\|_2\,\|\hat{\mathbf{e}}_T\|_2}.
\end{equation}
Since $\|\hat{\mathbf{e}}_S\|_2=\|\hat{\mathbf{e}}_T\|_2=1$, this further simplifies to $\mathcal{L}_{\mathrm{KD}} = 1 - \hat{\mathbf{e}}_S^{\top}\hat{\mathbf{e}}_T$.

\subsection{Graph Branch: Microenvironmental Modelling}
The graph branch captures cellular-level microenvironmental features as:

\noindent\textbf{1. Cell Segmentation and Feature Extraction.}
We employ NuLite for nucleus instance segmentation and type classification. For each detected cell $i$, we form a 51D node feature $\mathbf{h}_i^{(0)}$ by concatenating: (a) 37 morphology descriptors (e.g., region properties, Hu moments, and shape/boundary features, augmented with within-patch statistics), (b) a 6D one-hot vector of the predicted nucleus type, and (c) an 8D sinusoidal positional encoding of the centroid coordinates.

\noindent\textbf{2. Graph Construction.}
We build a patch-level undirected cell graph $\mathcal{G}=(\mathcal{V}, \mathcal{E})$ using nucleus centroids as spatial anchors. To reduce boundary artefacts, we discard nuclei near the patch border and mark patches with fewer than 5 remaining cells as invalid. We then construct a $k$-NN graph ($k=5$) by Euclidean centroid distance. Each edge $(i,j)$ is assigned an 18D attribute comprising a 16-kernel RBF embedding of inter-cell distance (cut-off 200 pixels) and a 2D unit direction vector from $j$ to $i$.

\noindent\textbf{3. Graph Representation Learning.}
We embed node and edge attributes into a shared 512D hidden space using lightweight MLP encoders, and apply a GNN backbone (GATv2 \cite{brody2022attentive}, TransformerConv \cite{vaswani2017attention}, or SAGPool \cite{zhang2023explainable}). We aggregate node embeddings into a graph-level representation via attentional readout (attention-based backbones) or hierarchical pooling (SAGPool), and map it with a projection head to obtain the graph embedding $\mathbf{v}_g \in \mathbb{R}^{D}$.

\noindent\textbf{4. Attention-based evidence extraction.}
For attention-based backbones (GATv2 \cite{brody2022attentive} and TransformerConv \cite{vaswani2017attention}), we take \emph{edge attention} as the last-layer attention coefficients $\alpha_{ij}$ (averaged over heads) and derive \emph{node attention} by aggregating incident $\alpha_{ij}$ and normalising within each patch. For SAGPool \cite{zhang2023explainable}, we use the pooling scores as node importance, but cannot report edge attentions.

\subsection{Visual Branch and Dual-branch Fusion Strategies}
\noindent\textbf{Visual Feature Extraction.}
ResNet-18 backbone processes the raw tile and produces a visual embedding $\mathbf{v}_v \in \mathbb{R}^{D}$ to complement GNN with tissue context.

\noindent\textbf{Fusion Strategies.} We fuse $\mathbf{v}_g$ and $\mathbf{v}_v$ to synergise cellular and contextual representations. In our implementation (and as summarised in Fig.~\ref{fig:framework}), we consider three lightweight fusion modules:

\noindent\textbf{1. Concatenation Fusion (Concat).} We can concatenate the two embeddings and pass them through an MLP layer, as follows:
\begin{equation}
    \mathbf{e}_S = \mathrm{MLP}_{\mathrm{cat}}\big([\mathbf{v}_v \,| |\, \mathbf{v}_g]\big).
\end{equation}

\noindent\textbf{2. Gated Fusion (Gate).}
We concatenate the embeddings and predict an element-wise gate $\boldsymbol{\lambda}\in(0,1)^D$ to balance the two branches:
\begin{equation}
\boldsymbol{\lambda}=\sigma\!\left(\mathrm{MLP}_{\mathrm{gate}}([\mathbf{v}_v\,||\,\mathbf{v}_g])\right),\quad
\mathbf{h}=\boldsymbol{\lambda}\odot\mathbf{v}_v+(1-\boldsymbol{\lambda})\odot\mathbf{v}_g,\quad
\mathbf{e}_S=\mathrm{MLP}_{\mathrm{out}}(\mathbf{h}).
\end{equation}
Here, $||$ denotes concatenation, and $\odot$ denotes the Hadamard product.

\noindent\textbf{3. Cross-Attention Interaction (Attn).} As illustrated in Fig.~\ref{fig:framework} ("Cross-attention interaction"), we retain patch-level image tokens from the visual encoder and use cell node embeddings to query them. Concretely, for each cell node embedding $\mathbf{h}_i$, we perform cross-attention with node queries and token keys/values, and update the node representation with a residual connection:
\begin{equation}
    \mathbf{h}_i \leftarrow \mathbf{h}_i + \mathrm{CrossAttn}(q=\mathbf{h}_i,\; k=v=\mathbf{T}),
\end{equation}
where $\mathbf{T}$ denotes the ResNet patch tokens. This allows each cell node to explicitly ``read'' visual evidence from its corresponding tile, improving distillation capacity and aligning graph-based explanations with the visual context.

\section{Results}

\begin{table}[!t]
    \centering
    \caption{Mean AUC over 5-fold cross-validation on Yale-HER2, SLN-Breast, and BRACS. \textbf{Bold} indicates the best, \textbf{\textit{bold italics}} the second best, and \underline{underline} the third best within each teacher group.}
    \label{tab:auc_by_teacher}
    \fontsize{8pt}{8.2pt}\selectfont

    \setlength{\tabcolsep}{1.0pt}      

    \begin{tabular}{@{}l c c c c c c c c c@{}}
        \toprule
        \multirow{2.5}{*}{\textbf{Method}} &
        \multicolumn{3}{c}{\textbf{Teacher: Phikon v2}} &
        \multicolumn{3}{c}{\textbf{Teacher: UNI v2}} &
        \multicolumn{3}{c}{\textbf{Teacher: Virchow2}} \\
        \cmidrule(lr){2-4} \cmidrule(lr){5-7} \cmidrule(lr){8-10}
        & \textbf{Yale} & \textbf{SLN} & \textbf{BRACS}
        & \textbf{Yale} & \textbf{SLN} & \textbf{BRACS}
        & \textbf{Yale} & \textbf{SLN} & \textbf{BRACS} \\
        \midrule

        PFM (Teacher)        & \textbf{0.927} & \textbf{0.950} & 0.783 & \textbf{0.937} & \textbf{0.956} & 0.790 & 0.855 & \textbf{0.935} & \textbf{0.846} \\

        \addlinespace[4pt]
        \multicolumn{10}{l}{\textit{--- Res18 Baselines ---}} \\
        w/o KD               & 0.791 & 0.803 & 0.702 & 0.791 & 0.803 & 0.702 & 0.791 & 0.803 & 0.702 \\
        Vanilla KD           & 0.879 & 0.853 & 0.783 & 0.876 & 0.857 & 0.793 & 0.843 & 0.862 & 0.801 \\

        \addlinespace[4pt]
        \multicolumn{10}{l}{\textit{--- Ours (Res18 + GNN) ---}} \\
        GATv2 (Concat)       & 0.910 & 0.867 & 0.784 & 0.897 & 0.901 & \textbf{0.806} & \textbf{\textit{0.894}} & 0.905 & 0.829 \\
        GATv2 (Gate)         & 0.920 & 0.864 & 0.787 & 0.913 & \underline{0.942} & 0.780 & 0.878 & 0.879 & 0.820 \\
        GATv2 (Attn)         & \textbf{\textit{0.924}} & 0.878 & \textbf{0.794} & 0.900 & 0.897 & 0.796 & 0.870 & \textbf{\textit{0.929}} & 0.833 \\

        SAGPool (Concat)     & 0.869 & 0.860 & \underline{0.789} & 0.913 & 0.899 & \textbf{\textit{0.805}} & \underline{0.884} & 0.888 & \textbf{\textit{0.839}} \\
        SAGPool (Gate)       & \underline{0.921} & \textbf{\textit{0.899}} & 0.785 & \textbf{\textit{0.917}} & \textbf{0.956} & 0.784 & 0.872 & 0.892 & 0.828 \\
        SAGPool (Attn)       & 0.911 & 0.868 & \textbf{\textit{0.792}} & \underline{0.916} & 0.906 & 0.797 & 0.858 & \underline{0.911} & \underline{0.836} \\

        TransGNN (Concat)    & 0.891 & \underline{0.883} & 0.775 & 0.890 & 0.933 & \underline{0.798} & 0.880 & 0.875 & \underline{0.836} \\
        TransGNN (Gate)      & 0.903 & 0.877 & 0.787 & \textbf{\textit{0.917}} & \textbf{\textit{0.943}} & 0.794 & \textbf{0.904} & 0.873 & 0.823 \\ 
        TransGNN (Attn)      & 0.910 & 0.863 & 0.785 & 0.905 & 0.938 & 0.803 & \textbf{0.904} & 0.875 & 0.821 \\

        \bottomrule
    \end{tabular}
\end{table}

\begin{figure}[!t]
    \centering
    \includegraphics[width=\textwidth]{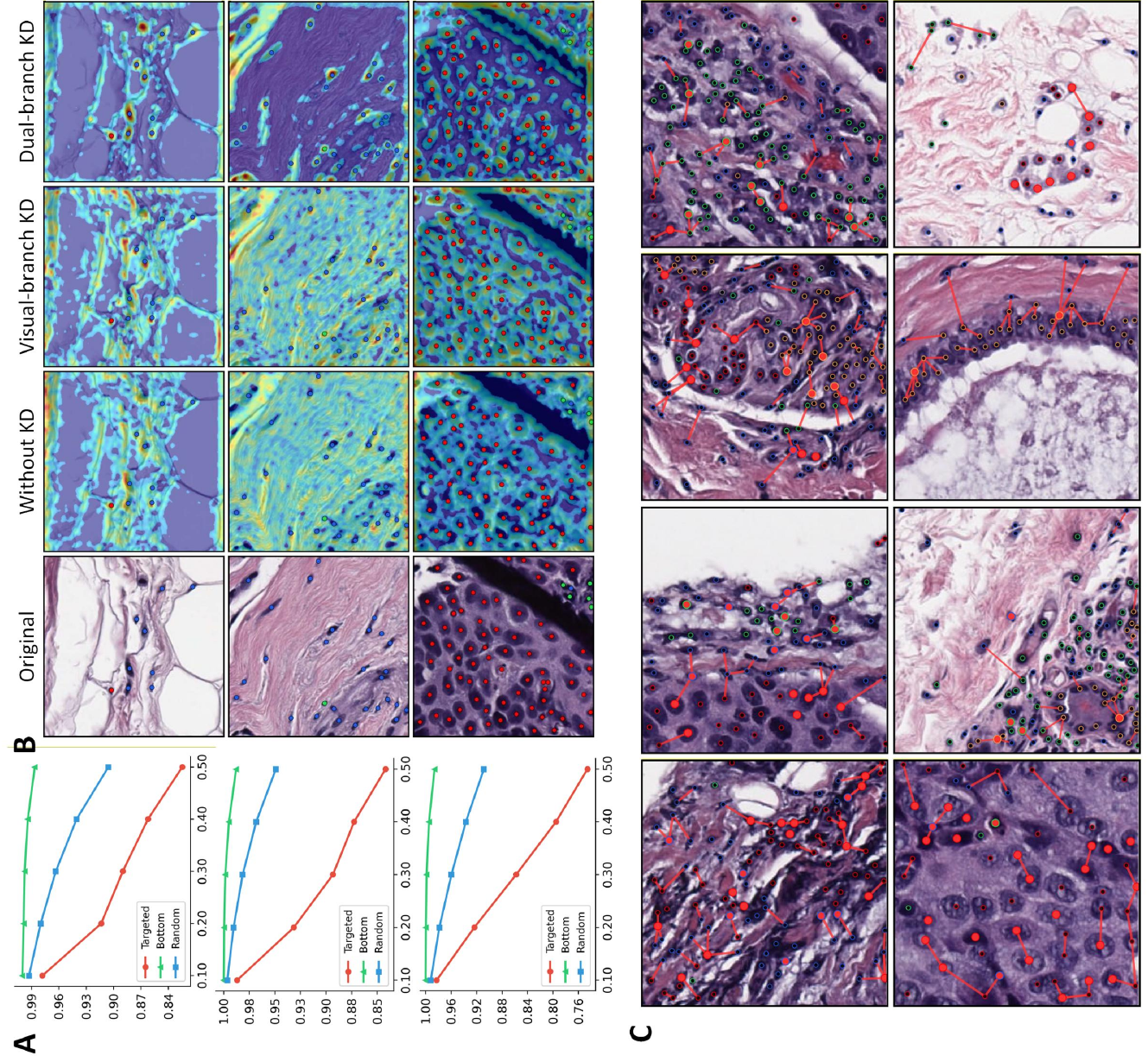} 
    \caption{\textbf{Faithfulness and visual refinement.} (A) Cosine similarity of teacher embeddings between the original tiles and tiles perturbed by occluding attention Top-$k$, Bottom-$k$, or random cells ($k$ refers to percentage). (B) Layer-1 Score-CAM maps for representative tiles from three ResNet-18 variants (ImageNet-pretrained, single-/dual-branch KD), with nuclei masks/overlays. (C) Tile-level graph evidence: node/edge attentions, with nodes of attention $>0.8$ highlighted (others shown in black); node borders denote cell types (Neoplastic/Red, Inflammatory/Green, Connective/Blue, Dead/Yellow, Epithelial/Orange); edges of attention $>0.8$ are shown in red.}
    \label{fig:faithfulness_refinement}
\end{figure}
\begin{figure}[t]
    \centering
    \includegraphics[width=\textwidth]{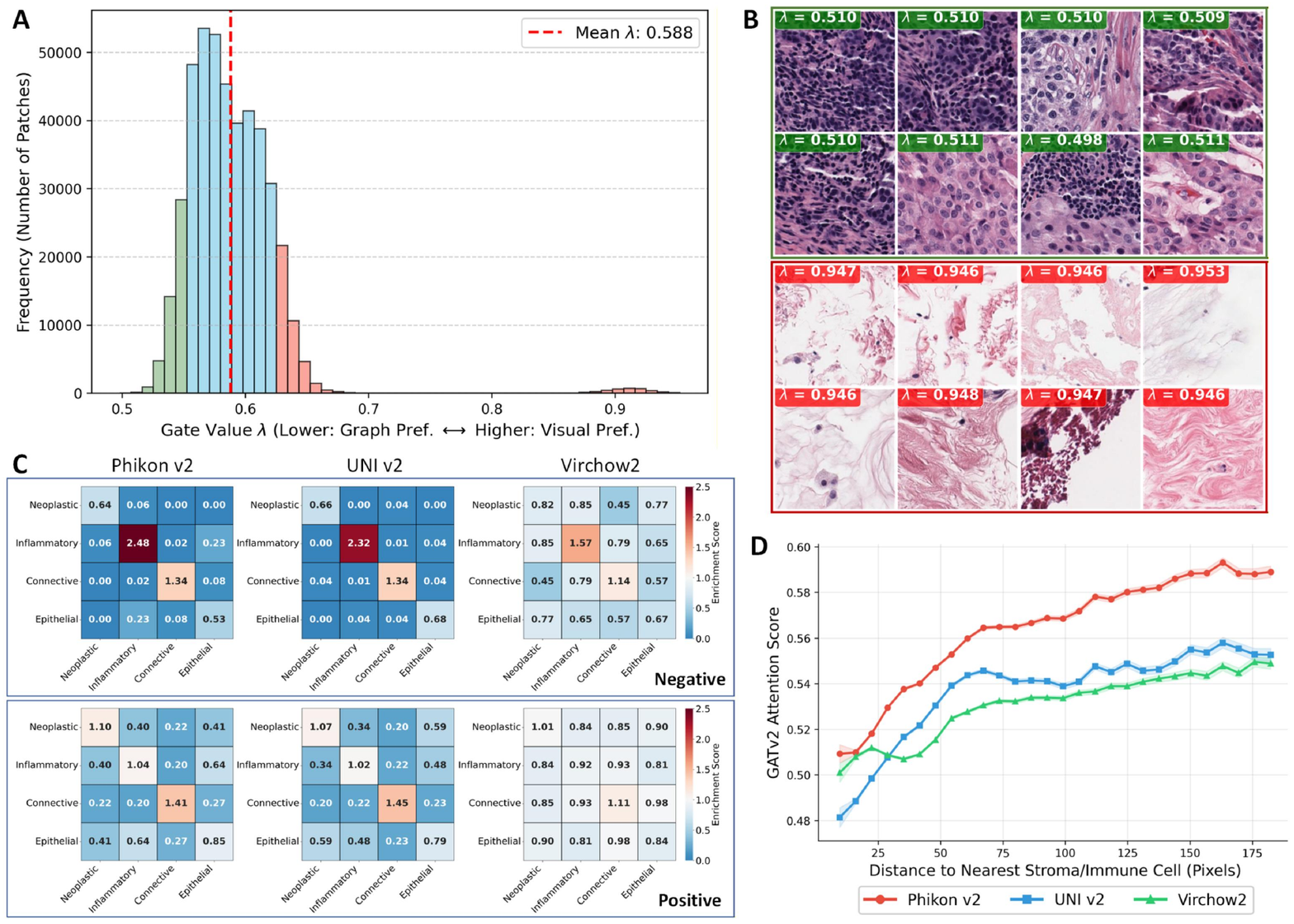} 
    \caption{\textbf{Adaptive Gating and Microenvironmental Analysis.} (A) Distribution of gate values $\lambda$ on Yale-HER2. (B) Example patches with low and high $\lambda$ values, with boxes highlighting regions. (C) Cell--cell communication enrichment heatmaps computed from the top 10\% highest-attention edges, shown separately for the \textit{Positive} and \textit{Negative} classes. (D) Scatter plot of nucleus-level attention scores versus the distance to the tumour--stroma boundary for neoplastic cells, with a fitted trend line.}
    \label{fig:gating_mechanism}
\end{figure}

\subsection{Datasets and Experimental Details}
We perform self-supervised distillation pre-training on TCGA-BRCA ($n=1{,}133$ slides from 40 sites). We evaluate generalisation on three external cohorts: Yale-HER2 \cite{farahmand2022deep,farahmand2022her2} ($n=192$), SLN-Breast \cite{campanella2019breast,campanella2019clinical} ($n=130$), and BRACS \cite{brancati2022bracs} ($n=547$). All slides are tiled at 20$\times$ into $256\times256$ patches. Yale-HER2 and SLN-Breast are binary classification benchmarks, while BRACS is a 7-class classification task.

For distillation, we sample cell-containing tiles and split them into 90\% training and 10\% validation. We train for 20 epochs with AdamW (lr $5{\times}10^{-4}$, wd $1{\times}10^{-5}$) and a 1-epoch linear warm-up, using a total batch size of 512 across four A800 GPUs; each epoch sub-samples 20\% of training tiles, and we select the checkpoint with the lowest validation loss. For downstream WSI classification, we extract frozen student tile embeddings and train an attention-based MIL aggregator (ABMIL \cite{ilse2018attention}) for 20 epochs (lr $1{\times}10^{-4}$) with 5-fold cross-validation, using final-epoch weights per fold for evaluation.

\subsection{Predictive Performance Evaluation}
We compare \textbf{G-Interp} with PFM teachers and the single-branch ResNet-18 KD baseline (Vanilla KD) across the three cohorts (Table~\ref{tab:auc_by_teacher}). G-Interp consistently improves the area under the receiver operating characteristic curve (AUC) over the single-branch baseline, including on BRACS (7-class classification). Despite reducing the parameter count by over $20\times$, it matches or exceeds the teacher in settings, indicating that explicit microenvironment modelling can regularise distillation and suppress task-irrelevant variation in teacher embeddings. Across fusion variants, no single mechanism is consistently dominant.

\subsection{Interpretability and Biological Faithfulness}
Beyond predictive performance, \textbf{G-Interp} provides microenvironmental interpretability by revealing cell- and interaction-level evidence for PFM decisions. We assess whether the student highlights biologically meaningful evidence, focusing on: (1) cellular faithfulness, (2) adaptive gate behaviour, and (3) microenvironment interaction patterns. Qualitative visualisations use the \textit{GATv2+Gate} student on Yale-HER2.

\textbf{Cellular Faithfulness and Visual Refinement.}
We test whether high-attention cells identified by the graph branch are causally important by synchronising occlusions across both branches and re-measuring similarity to the frozen teacher. As shown in Fig.~\ref{fig:faithfulness_refinement}(A), occluding Top-$k$ cells (highest GAT attention) consistently causes a larger drop in cosine similarity of the tile embeddings than occluding Bottom-$k$ or random cells, across all three teachers. We visualise the corresponding salient nodes and edges in Fig.~\ref{fig:faithfulness_refinement}(C).
We further examine whether joint distillation refines the visual encoder. Layer-1 Score-CAM maps (Fig.~\ref{fig:faithfulness_refinement}B) show that the dual-branch student concentrates activation around nuclei more tightly than the ImageNet-pretrained and single-branch KD baselines, suggesting that the graph prior suppresses background responses.

\textbf{Adaptive Gating Mechanism.}
We analyse how the gate value $\lambda$ controls the relative contribution of the two branches, where low $\lambda$ emphasises the graph embedding and high $\lambda$ emphasises the visual embedding. Fig.~\ref{fig:gating_mechanism}(A) shows that $\lambda$ spans a wide range across tiles, indicating that the model does not use a fixed mixing ratio but adjusts the fusion weights across samples. Example tiles (Fig.~\ref{fig:gating_mechanism}(B)) suggest that low $\lambda$ is often associated with cell-dense regions with richer local interactions, whereas high $\lambda$ tends to occur in cell-sparse background regions (e.g., stroma/adipose).

\textbf{Microenvironment Interaction Patterns.}
We aggregate the top 10\% highest-attention edges to form class-wise cell--cell communication heatmaps (Fig.~\ref{fig:gating_mechanism}(C)). Here, \emph{same-type} interactions refer to edges linking two cells of the same predicted type, while \emph{cross-type} interactions link different cell types. While same-type interactions are overall dominant, the \textit{Positive} class shows relatively stronger cross-type connectivity. Notably, the Virchow2-distilled model exhibits comparatively increased cross-type interactions even in the \textit{Negative} class, suggesting a stronger preference for diverse inter-cellular relations.
To quantify spatial patterns, we relate neoplastic-cell attention to the distance from the tumour--stroma boundary (Fig.~\ref{fig:gating_mechanism}(D)). Across teachers, attention is higher for cells deeper within tumour nests; compared with UNI v2 and Phikon v2, Virchow2 shows a flatter decay with distance, indicating a weaker contrast between boundary-adjacent and deep-tumour neoplastic cells.

\section{Conclusion}
We presented \textbf{G-Interp}, a graph-visual distillation framework that adapts frozen pathology foundation models into lightweight students while providing cellular- and interaction-level evidence. By distilling teacher embeddings into a dual-branch student with a microenvironment graph interpreter and adaptive fusion, G-Interp achieves strong performance across three external breast cancer cohorts. Beyond AUC, we showed that graph attention supports faithful cell-level attribution, the gating mechanism responds to tissue content, and the distilled students reveal teacher-specific microenvironmental preferences in how attention varies across cell types and spatial contexts. These results position G-Interp as a practical route to efficient deployment and a tool for analysing how PFMs encode microenvironmental signals.

Future work will extend G-Interp to broader cancer types and evaluate robustness under varied staining and scanning conditions.

\begin{credits}
\subsubsection{\ackname} This work was supported by the following grants. JR is supported by the NIHR Oxford Biomedical Research Centre. KY is supported by the National Natural Science Foundation of China (No. 62476101) and the Guangdong Basic and Applied Basic Research Foundation (Grant No. 2024A1515140137). 
The views expressed are those of the authors and not necessarily those of the NHS, the NIHR, or the Department of Health.

\subsubsection{\discintname}
The authors have no competing interests to declare that are relevant to the content of this article.
\end{credits}

\newpage

\bibliographystyle{splncs04}
\bibliography{paper-0004}


\end{document}